\documentclass{article}
\usepackage[final]{mlncp_2023}
\usepackage{url}
\usepackage{graphicx} 

\usepackage{color}
\usepackage{multirow}

\newcommand{\mozhgan}[1]{\textcolor{black}{#1}}

\title{Squeezed Edge YOLO: Onboard Object Detection on Edge Devices}

\author{Edward Humes$^1$ \quad Mozhgan Navardi$^2$ \quad Tinoosh Mohsenin$^2$ \\
$^1$University of Maryland, Baltimore County \quad $^2$Johns Hopkins University \\
\texttt{ehumes2@umbc.edu} \quad
\texttt{\{mnavard1, tinoosh\}@jhu.edu}\\
}
\date{October 2023}

\begin{document}

\maketitle

\begin{abstract}

\mozhgan{Demand for efficient onboard object detection is increasing due to its key role in autonomous navigation. However, deploying object detection models such as YOLO on resource constrained edge devices is challenging due to the high computational requirements of such models. In this paper, 
an compressed object detection model named Squeezed Edge YOLO is examined. This model is compressed and optimized to kilobytes of parameters in order to fit onboard such edge devices. To evaluate Squeezed Edge YOLO, two use cases - human and shape detection - are used to show the model accuracy and performance. Moreover, the model is deployed onboard a GAP8 processor with 8 RISC-V cores and an NVIDIA Jetson Nano with 4GB of memory. Experimental results show Squeezed Edge YOLO model size is optimized by a factor of 8x which leads to 76\% improvements in energy efficiency and 3.3x faster throughout.}

\end{abstract}

\section{Introduction}
Interest in Machine Learning~(ML) is dramatically increasing as it provides a promising solution for various applications such as autonomous navigation~\cite{rausch2017learning, kuutti2020survey}. Object detection models in particular can significantly assist in autonomous navigation by detecting obstacles and pre-defined objects of interest in the environment~\cite{lamberti2023bio}. However, object detectors have high computational requirements due to the need for accuracy and the ability to detect various object categories. GPUs with significant computational capacity are often mandatory to train such complex models, yet onboard processing and edge computing necessitates low-power and low-computation algorithms as a result of the limited power and computational capacity available~\cite{mazumder2023reg}.

Object detectors are trained classifiers that can identify and locate multiple objects within an image. These detectors are trained on a set of annotated images, and their accuracy is evaluated on unseen datasets. There are two commonly used object detector paradigms: single-shot and two-shot. Single-shot-based methods such as You Only Look Once~(YOLO)~\cite{yolo2022}, Single Shot Detector~(SSD)~\cite{ssd2016}, etc., directly predict the class probabilities and Bounding Box~(BBox) coordinates for objects in an image. In contrast, two-shot architectures such as R-CNN~\cite{rcnn2014}, Faster R-CNN~\cite{fasterrcnn2015}, etc., generate a set of region proposals and then classify and refine them to output the final object detection. Moreover, two-shot object detection methods have several advantages over other methods, including robustness to scale and size variations, accurate localization, flexibility, and improved object recognition~\cite{aaai2023tejaswini}. 
However, these advantages come at the expense of inference speed, with single-shot object detectors generally being faster than two-shot object detectors. Despite this, even single-shot objector models are difficult to deploy to resource constrained edge devices due to their high computational complexity. Therefore, it is important to improve object detection models to meet power consumption and real-time requirements on such devices~\cite{edgeyolo2022, arnabmicro}.

In recent years, researchers have presented optimized object detection models~\cite{edgeyolo2022, yololite2018, edgecomm2022mozhgan, loquercio2018dronet, palossi201964, micro2023mozhgan, moosmann2023tinyissimoyolo, lamberti2021low} to enable onboard object detection on edge devices. Work in~\cite{edgecomm2022mozhgan, loquercio2018dronet, palossi201964} proposed an optimized model for collision avoidance and obstacle detection to enable autonomous navigation for a tiny drone with a GAP8 processor. An Automatic License Place Recognition~(ALPR) is proposed on~\cite{lamberti2021low} which leverages a GAP8 processor. TinyissimoYOLO is proposed in~\cite{moosmann2023tinyissimoyolo} which is deployed on different microcontrollers, achieving an accuracy of 58.5\% for three objects. In~\cite{edgeyolo2022}, an optimized YOLO model, EdgeYOLO, is presented that achieves a reduction in model size by about 10x from around 250~MB to 25~MB. They deployed their proposed model on an NVIDIA Jetson board for system-level verification.  In~\cite{micro2023mozhgan} a metareasoning framework is proposed for an efficient goal-oriented autonomous navigation by switching between an optimized YOLO model, Squeezed Edge YOLO, and a lighter Convolutional Neural Network~(CNN) based model. 

In this work, the focus is on the Squeezed Edge YOLO~\cite{micro2023mozhgan} with detailed implementation and more experimental results. Then, we compare the model with the state-of-the-art work in~\cite{edgeyolo2022} which have similar accuracy when we deployed both models on a JetBot equipped with an NVIDIA Jetson Nano board to show the latency and energy improvement.
Moreover, we deploy the Squeezed Edge YOLO model on extremely resource constrained devices such as the GAP8 processor while meeting real-time requirements. The main contribution of this work is as follows:
\begin{itemize}

\item A detailed implementation of an energy efficient Squeezed Edge YOLO as a tiny Machine Learning~(tinyML) model is provided in this work for tiny edge devices.
\item Onboard YOLO-based model deployment and edge computing on resource constrained edge devices such as the GAP8 and NVIDIA Jetson Nano.
\item Power and latency analysis on GAP8 processor and NVIDIA Jetson Nano as a result of the memory hierarchy and onboard data transmission overhead.

\end{itemize}\vspace*{3pt}

To evaluate the Squeezed Edge YOLO\cite{micro2023mozhgan}, we deployed 
it on the GAP8 processor. Moreover, we compared the throughput and energy efficiency of the model on an NVIDIA Jetson Nano board with state-of-the-art-works. Experimental results show a 3.3x faster throughput while achieving a 76\% energy efficiency improvement.


\section{Proposed Approach}
\label{SEC:PropApp}

In this section, we describe challenges in deploying object detection neural networks to resource-constraint edge devices as a motivation. Then, we propose a hardware-aware optimized object detection model named Squeezed Edge YOLO for onboard processing.


\subsection{Motivation}
While we were able to deploy EdgeYOLO~\cite{edgeyolo2022} to the Jetson Nano, our attempt to deploy the base EdgeYOLO model onboard the GAP8 resulted in failure. Due to the model's size, it would overflow the limited stack space available on the GAP8 and crash. Therefore, we applied hardware-aware model optimizations by profiling EdgeYOLO, and working to reduce memory consumption while attempting to retain an acceptable level of inference accuracy. For this aim, in the next section we discussed relevant hardware features and presented an optimized Squeezed Edge YOLO.


\subsection{GAP8 Hardware Architecture}

\begin{figure*}
\centerline{\includegraphics[width=30pc]{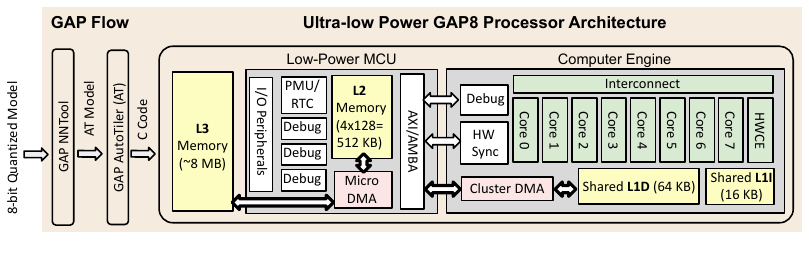}}
\caption{GAP8 processor architecture and custom GreenWaves toolchain. As the figure shows, the GAP8 has a variety of hardware features in order to make the most of its limited power budget and be able to run neural networks on the edge~\cite{flamand2018gap, gap8gw, micro2023mozhgan}.}\label{fig1}  
\end{figure*}

The GAP8 is a microcontroller designed specifically with tinyML in mind~\cite{flamand2018gap}, and as shown in Figure~\ref{fig1}, it contains one main processor core capable of running at up to 250 MHz, and an octacore cluster capable of running at up to 175 MHz. In addition, it contains a variety of hardware accelerators including a convolution engine, as well as a 3-layer memory hierarchy composed of 64KB of L1 RAM, 512 KB of L2 RAM, and an off-chip L3 memory composed of many megabytes of RAM and flash. Past work with the GAP8 has demonstrated that it is primarily constrained by its memory hierarchy~\cite{umbcreview}, rather than its computational capabilities. Most modern microprocessors go to great lengths to mask memory latency, such as by having multiple layers of caches on top of RAM, hardware-prefetching, out of order execution, etc. The GAP8's memory hierarchy is entirely software-managed, L1 memory is the fastest, needing only a single cycle to access, L2 memory takes multiple cycles to access, while L3 memory is the slowest - it is not memory mapped, requiring Direct Memory Access~(DMA) transfers from the off-chip L3 RAM into L2 memory in order to read and modify data stored within. Thus, it is the responsibility of applications to arrange their data in an optimal layout around the memory hierarchy to minimize stalls by the GAP8's in-order processor cores and accelerators. 

\begin{table}
\centering
\makebox[0pt][c]
{\parbox{\textwidth}{%
    \begin{minipage}[b]{0.47\hsize}\centering
        \caption{Backbone feature extraction~\cite{micro2023mozhgan}}
        \label{tab:backbone}
        
        \resizebox{0.975\columnwidth}{!}{%
        
        \begin{tabular}{ccccc}
\hline 
Layer & Type    & Size/Stride         & Filters         & Output   \\ \hline
0     & Conv    & 3x3/2               & 16              & 64x64x16 \\
1     & Conv    & 3x3/2               & 32              & 32x32x32 \\
2     & SE      & -                   & -               & 32x32x32 \\
3     & Conv    & 3x3/1               & 16              & 32x32x16 \\
4     & SE      & -                   & -               & 32x32x16 \\
5     & Conv    & 3x3/1               & 16              & 32x32x16 \\
6     & SE      & -                   & -               & 32x32x16 \\
7     & Route   & \multicolumn{2}{c}{Input: {[}6, 4{]}} & 32x32x32 \\
8     & Conv    & 3x3/1               & 32              & 32x32x32 \\
9     & SE      & -                   & -               & 32x32x32 \\
10    & Route   & \multicolumn{2}{c}{Input: {[}2, 9{]}} & 32x32x64 \\
11    & SE      & -                   & -               & 32x32x64 \\
12    & MaxPool & 2x2/2               &                 & 16x16x64 \\ \hline
\multicolumn{5}{c}{} \\  
\multicolumn{5}{c}{} \\  
\multicolumn{5}{c}{} \\  
\multicolumn{5}{c}{} \\  
\multicolumn{5}{c}{} \\  
\end{tabular} }
        
    \end{minipage}
    \hspace{0.01\hsize}
    \begin{minipage}[b]{0.47\hsize}\centering
        \caption{Neck feature enhancement~\cite{micro2023mozhgan}}
        \label{tab:neck}
        \resizebox{\columnwidth}{!}{%
        \begin{tabular}{ccccc}
\hline 
Layer & Type     & Size/Stride          & Filters         & Output    \\ \hline
0     & Conv     & 3x3/1                & 64              & 16x16x64  \\
1     & Conv     & 3x3/2                & 128             & 8x8x128   \\
2     & SE       & -                    & -               & 8x8x128   \\
3     & Conv     & 1x1/1                & 256             & 8x8x256   \\
4     & SE       & -                    & -               & 8x8x256   \\
5     & Conv     & 1x1/1                & 256             & 8x8x256   \\
6     & Conv     & 3x3/1                & 128             & 8x8x128   \\
7     & Route    & \multicolumn{2}{c}{Input: {[}5,{]}}    & 8x8x256   \\
8     & Conv     & 3x3/1                & 64              & 8x8x128   \\
9     & Upsample & 2x2                  & -               & 16x16x128 \\
10    & Conv     & 1x1/1                & 64              & 16x16x64  \\
11    & Route    & \multicolumn{2}{c}{Input: {[}10, 0{]}} & 16x16x128 \\
12    & Conv     & 3x3/1                & 128             & 16x16x128 \\
13    & SE       & -                    & -               & 16x16x128 \\
14    & Conv     & 1x1/1                & 128             & 16x16x128 \\
15    & SE       & -                    & -               & 16x16x128 \\
16    & Conv     & 3x3/1                & 64              & 16x16x64  \\
17    & Detect   & \multicolumn{2}{c}{Input: {[}6, 16{]}} &           \\ \hline
\end{tabular}}

    \end{minipage}
    
}}
\vspace*{-15pt}
\end{table}

\subsection{Squeezed Edge YOLO}
Given this information, our first step was to shrink EdgeYOLO's original input image size specifically when running on the GAP8. 
As the AI-deck expansion board (which includes the GAP8 processor) 
can only capture images at a resolution of 324x244, we shrunk the input image size from 416x416x3 to 128x128x3. Besides this though, 128x128x3 images only require 49152 bytes, or less than the whole of L1 memory. In practice, the image will almost never be stored in its entirety within L1 RAM - network parameters also need to be stored in L1 memory for efficient retrieval, although the smaller size does make moving various data tiles much easier. We briefly considered using a higher image resolution, albeit in gray scale, for object detection, however we ended up deciding against this, given that we wanted to identify specific colored objects for reinforcement learning purposes. Fortunately, the smaller size still manages to achieve a good trade-off between inference latency and accuracy for object detection. 

More broadly, besides input image size, we focused our attention on other aspects contributing to high memory usage, which more often than not was a combination of convolutional layers and residual blocks. By default, the base Edge YOLO model produces many output channels for each convolutional layer, while this assists feature extraction in images, these output channels must be stored somewhere. Besides the large number of channels, another challenge for the GAP8 with these convolutional layers is the size of the weights required. While 1x1 convolutions are not as great for extracting features present within an image, they require 9x less weights than a 3x3 convolution. Thus, as depicted in Table~\ref{tab:backbone}~and~\ref{tab:neck}, where possible, we attempted to either reduce the number of input and output channels present in EdgeYOLO convolutional layers, or if that was still not enough, switch to using 1x1 convolutional layers. 

Residual blocks~\cite{resnet2016} are quite common in modern computer vision networks, as they both ameliorate the problem of vanishing gradients, and reduce total parameter count. However, despite using less parameters than a network with many layers stacked over each other, they are taxing on the GAP8's memory hierarchy. This is because in addition to the weights and activations of the current neural network layer being operated on, intermediate activations in skip connections need to be stored somewhere in memory until they are fed into another layer. The base EdgeYOLO model, and most modern YOLO implementations generally speaking, often contain deeply nested skip connections covering dozens of layers, some of which contain their own skip connections. Shuffling these around the memory hierarchy in tandem with the regular execution of the neural network is challenging, and so we removed many residual blocks, and simplified others to only covering a few layers. In addition, we removed one of the three detection heads; we figured that as the images we were operating on were already quite small, and the detection heads are primarily for performing object detection at different scales, it would be somewhat redundant. By extension, this also removed the longest skip connection, along with its associated inner skip connections.

\begin{figure*}
\centerline{\includegraphics[width=30pc]{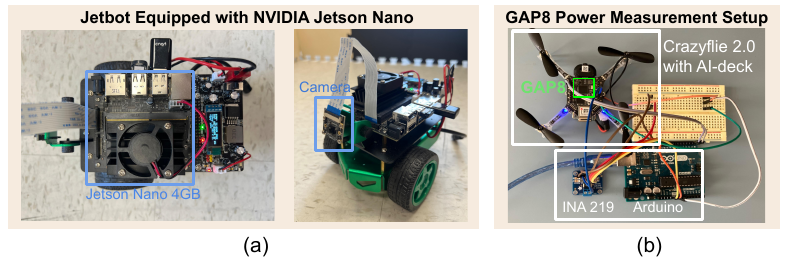}}
\caption{A JetBot used for experimentation as well as a power measurement setup for the GAP8. (a) The JetBot contains a Jetson Nano as well as various peripherals including a camera.
(b) The INA 219 and Arduino are used in this paper to measure the GAP8 processor power consumption when the model is deployed onboard.}\label{fig2}  \vspace*{-5pt}
\end{figure*}

\section{Experimental Results}
\label{SEC:Expr}

In this section, we describe our evaluation of Squeezed Edge YOLO deployed on two target edge devices: an NVIDIA Jetson Nano with 4GB of RAM shared between the CPU and GPU, and an AI-deck with a GAP8 microcontroller onboard. Moreover, we specifically focus on: inference latency on each processor, power consumption, and inference accuracy in comparison with the state-of-the-art-work in~\cite{edgeyolo2022}.
\subsection{Experimental Setup}
For experiments, we used a JetBot with an NVIDIA Jetson Nano containing 4GB of RAM as shown in Figure~\ref{fig2}~(a), as well as a Crazyflie drone with an AI-deck expansion board, which contains a GAP8 microcontroller. The Jetson Nano contains hardware onboard for measuring the power draw of the Tegra X1's CPU and GPU cores. Meanwhile, the GAP8 required the use of external hardware. Figure~\ref{fig2}~(b) shows an INA219 and an Arduino connected to the AI-deck's power pins in order to measure average power usage during the runtime of the network. As we were primarily interested in autonomous navigation, we created a custom dataset for testing, composed of 1500 images of various shapes captured from our deployment platforms. We additionally created a separate mixed dataset of human images using 3500 images taken from CrowdHuman~\cite{shao2018crowdhuman}, and 3500 more images we captured using the AI-deck.

\subsection{Model Optimization and Onboard Object Detection}

\begin{figure}
\centerline{\includegraphics[width=34pc]{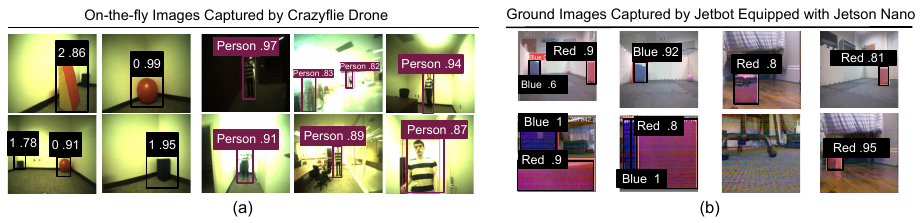}}
\caption{(a)~Squeezed Edge YOLO object detection images (first two column), along with human detection images (last three column) captured using the Crazyflie's AI-deck expansion board. (b)~
Object detection images captured using the RGB camera on the JetBot. The overlaid labels show the classification and detected accuracy by Squeezed Edge YOLO model.}\label{fig3}  \vspace*{-5pt}
\end{figure}

Once we had our final object detection model, it achieved an acceptable inference latency for our purposes aboard the AI-deck: roughly 8 inferences a second, displayed in Table~\ref{tab3}, and as shown in Figure~\ref{fig3} achieving an acceptable level of accuracy for object detection tasks. Upon examining hardware utilization in the cycle-accurate GAP8 simulator, named GVSOC~\cite{gvsoc}, we found it had relatively low micro-DMA usage as a percentage of total runtime, as well as few CPU stalls on the cluster cores. This indicates that the memory hierarchy is being used effectively, the cores are being kept busy with crunching through the neural network, as the data is where it needs to be when the processor begins operating on it.

\subsection{Onboard Model Implementation}
In this section, we describe the steps involved in deploying both EdgeYOLO and the Squeezed Edge YOLO model we developed to the target hardware platforms, the GAP8 and Jetson Nano. Additionally, the process of extracting power and latency measurements is described.

\begin{figure*}
\centerline{\includegraphics[width=30pc]{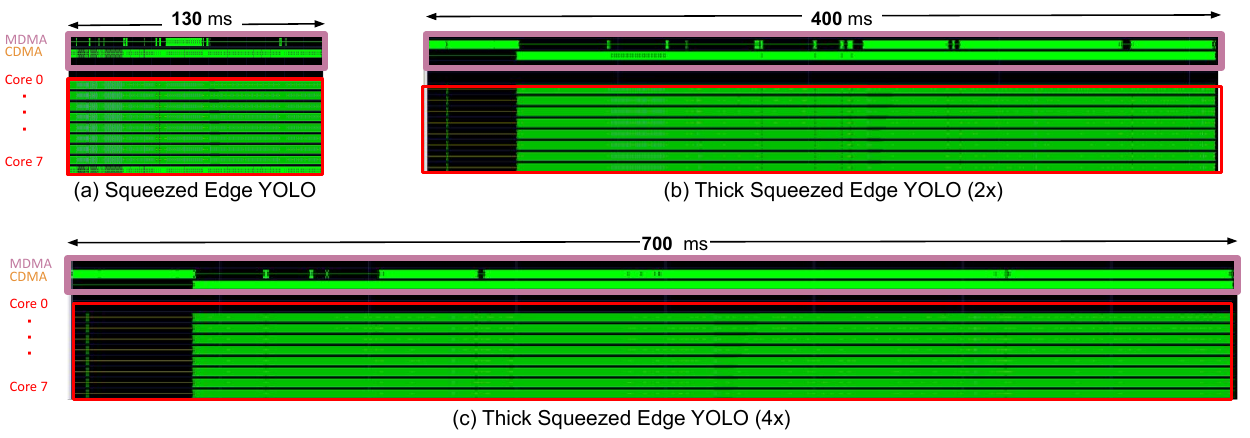}}
\caption{Inference phase recorded in VCD traces for different size of Squeezed Edge YOLO model on GAP8: (a)~Squeezed Edge YOLO~\cite{micro2023mozhgan} (b)~an enlarged Squeezed Edge YOLO (2x parameters) (c)~an enlarged Squeezed Edge YOLO (4x parameters). The green portion represents hardware utilization. Cluster-DMA~(CDMA) and Micro-DMA~(MDMA) are responsible for data transfer between L1, L2 and L3 memories in GAP8, respectively.}\label{fig4}  \vspace*{-6pt}
\end{figure*}

\textbf{GAP8 Processor.}
GreenWaves Technologies~\cite{flamand2018gap}, the company that developed the GAP8, publishes a custom toolchain~\cite{greenwaves} to ease deploying neural networks to the GAP8. This toolchain, depicted in Figure~\ref{fig4} consists of a modified GCC compiler with GAP8 support added in, the Neural Network Tool (NNTOOL), and AutoTiler. NNTOOL is responsible for converting adjusting ONNX and TFLite models to replace unsupported operations, as well as to export the weights into a ROM image. AutoTiler, takes care of converting the models execution graph into C code making use of optimized compute kernels, as well as tiling to manage memory transfers for memory locality.

\begin{table*}[]
\centering
\caption{For a fair comparison, models are re-implemented. The size, power consumption, and inference latency of YOLOv5s~\cite{aaai2023tejaswini}, EdgeYOLO~\cite{edgeyolo2022}, and the Squeezed Edge YOLO~\cite{micro2023mozhgan} is extracted. Squeezed Edge YOLO is successfully deployed to the GAP8 processor which is about 8x smaller than YOLOv5s. The shapes dataset is used for this experiments. \newline} \label{tab3}
\resizebox{\textwidth}{!}{%

\begin{tabular}{|l|c|c|c|cc|}
 \hline
Model              & Parameters 
& Accuracy & Model Size      & \multicolumn{1}{c|}{Power}      & Inference Latency      \\ \hline
YOLOv5s~\cite{aaai2023tejaswini}     & 7.3 M        
&    0.96 mAP      & 237 Mb (32-bit) & \multicolumn{1}{c|}{N/A}        & N/A                    \\ \hline
EdgeYOLO~\cite{edgeyolo2022}   & 8.1 M      
&     0.99 mAP     & 65 Mb (8-bit)   & \multicolumn{2}{c|}{Failed due to memory limitations} \\ \hline
\textbf{Squeezed Edge YOLO~\cite{micro2023mozhgan} }& \textbf{931 K ($\sim$8x smaller)}     
&   \textbf{ 0.95 mAP}      & \textbf{7.5 Mb (8-bit)}   & \multicolumn{1}{c|}{\textbf{541 mW}}           & \textbf{130 ms   }              \\ \hline
\end{tabular}}\vspace*{-12pt}
\end{table*}

The GAP8 does not contain any floating point units~\cite{umbcreview}, and so we quantized the Squeezed Edge YOLO model to 8-bit integers, which also has the side effect of reducing RAM usage. We then used GreenWaves' toolchain to convert the model into C code which we then compiled and benchmarked both in GVSOC and on the AI-deck. As GVSOC is designed to be a cycle-accurate emulator of the GAP8, we were able to extract performance info, as well as hardware utilization in the form of VCD traces, whereas deployment to the AI-deck allowed extracting energy utilization. The VCD traces found in Figure~\ref{fig4} showed that Squeezed Edge YOLO made efficient use of the memory hierarchy. 
While the much larger, and much slower Squeezed Edge YOLO models that we designed, made heavy use of L3 RAM.

\textbf{NVIDIA Jetson Nano 4GB.}
To show the efficiency of the Squeezed Edge YOLO in comparison with the proposed model in~\cite{edgeyolo2022}, we deployed both models on the NVIDIA Jetson Nano as EdgeYOLO does not meet GAP8 resource constraints. The Jetson Nano is much closer to the workstations and servers typically used for training and deploying machine learning models, given that it has an ARMv8 CPU, and NVIDIA Maxwell architecture GPU with CUDA support, along with supporting both FP16 and FP32 representations. Thus, it can directly execute PyTorch and TensorFlow models on both its CPU and GPU. However, NVIDIA provides a machine learning framework named TensorRT~\cite{tensorrt}, which applies GPU-specific optimizations in order to boost the performance of models during execution. Thus, we converted our models to TensorRT before executing them on-device. Additionally, as the Jetson Nano does not face the same memory constraints as the GAP8, we tested all models aboard the Jetson Nano using a 3 channel 416x416 input image size. The Jetson Nano is able to monitor power going to both the CPU and GPU respectively, obviating the need for the same setup we used for measuring power on the GAP8 as reported in Table~\ref{tab4}. Experimental results show Squeezed Edge YOLO used less CPU and GPU power consumption in comparison EdgeYOLO~\cite{edgeyolo2022}.

\begin{table}[t]
\caption{Hardware implementation results that compare Squeezed Edge YOLO with the work presented in~\cite{edgeyolo2022}. As EdgeYOLO was unable to be deployed to the GAP8, we used the Jetson Nano as our baseline hardware for comparison purposes.}
\label{tab4}
\begin{center}
{
	\begin{tabular}{ |c|c|c|}
\hline 
  Metric/Approach & EdgeYOLO~\cite{edgeyolo2022} & Squeezed Edge YOLO~\cite{micro2023mozhgan} \\
\hline 
	Inference Latency (ms)    &  231   &  70\\
\hline 	
    Throughput (Inference/Sec)       & 4.2 & \textbf{14.2 (3.3x faster)}\\
	
\hline 	
    CPU Power Consumption (mW)  & 777 &  1158\\
\hline 	
    GPU Power Consumption (mW)  & 3846 &  2434\\

\hline 	
    Performance (GOPS)  & 198.2 &   94.3\\
\hline 	
    Energy Efficiency (GOPS/J)  & 185.7 &  \textbf{788.5} \\
    
\hline 	
   Energy/Inference (mJ)  & 1068 &  \textbf{251.5 (76\% improvement)} \\
\hline

    \end{tabular}
    }
\end{center}
\vspace{-18pt}
\end{table}

\section{Conclusion}
\label{SEC:Conc}
\mozhgan{
In this work, we examined an optimized YOLO model named Squeezed Edge YOLO for onboard processing on resource constrained edge devices containing limited sources of power and computational capacity. In order to achieve this, a baseline YOLO model was optimized in terms of its number of parameters and computations to reduce the model size. Evaluation was carried out on a drone known as the Crazyflie with a GAP8 processor containing 8 RISC-V cores, and a JetBot with an NVIDIA Jetson Nano containing 4 GB of memory. The experimental results show that Squeezed Edge YOLO is about 8x smaller than existing edge focused YOLO model. We measured power consumption and latency upon deploying Squeezed Edge YOLO to the target devices, and found that by optimizing the YOLO model, it achieves a 76\% and 3.3x energy efficiency and throughput improvement on the Jetson Nano board.}

\acksection
This project was sponsored by the U.S. Army Research Laboratory under Cooperative Agreement Number W911NF2120076.

\bibliographystyle{unsrt}
\bibliography{ref.bib}

\end{document}